# Building chatbots from large scale domain-specific knowledge bases: challenges and opportunities

Walid Shalaby, Adriano Arantes, Teresa GonzalezDiaz, Chetan Gupta
Hitachi America Ltd.
Santa Clara, CA, USA
{walid.shalaby, adriano.arantes, teresa.gonzalezdiaz, chetan.gupta}@hal.hitachi.com

## ABSTRACT

Popular conversational agents frameworks such as Alexa Skills Kit (ASK) and Google Actions (gActions) offer unprecedented opportunities for facilitating the development and deployment of voice-enabled AI solutions in various verticals. Nevertheless, understanding user utterances with high accuracy remains a challenging task with these frameworks. Particularly, when building chatbots with large volume of domain-specific entities. In this paper, we describe the challenges and lessons learned from building a large scale virtual assistant for understanding and responding to equipment-related complaints. In the process, we describe an alternative scalable framework for: 1) extracting the knowledge about equipment components and their associated problem entities from short texts, and 2) learning to identify such entities in user utterances. We show through evaluation on a real dataset that the proposed framework, compared to off-the-shelf popular ones, scales better with large volume of entities being up to 30% more accurate, and is more effective in understanding user utterances with domain-specific entities.

## 1 Introduction

Virtual assistants are transforming how consumers and businesses interact with Artificial Intelligence (AI) technologies. They provide users with a natural language interface to the backend AI services, allowing them to ask questions or issue commands using their voice. Nowadays, voice-activated AI services are ubiquitously available across desktops, smart home devices, mobile devices, wearables, and much more, helping people to search for information, organize their schedule, manage their activities, and accomplish many other day-to-day tasks.

The use of digital assistants technology, though widely adopted in the consumer space, haven't seen the same degree of interest in industrial and enterprise scenarios. One key challenge for the success of these voice assistants is the so-called conversational Language Understanding (LU), which refers to the agent's ability to understand user speech through precisely identifying her intent and extracting elements of interest (entities) in her utterances. Building high quality LU systems requires: 1) lexical knowledge, meaning access to general, common-sense, as well as domain-specific knowledge related to the task and involved entities, and 2) syntactic/semantic knowledge, meaning "ideally" complete coverage of all different ways users might utter their commands and questions.

Several conversational agents frameworks were introduced in recent years to facilitate and standardize building and deploying voice-enabled personal assistants. Examples include Alexa Skills Kit (ASK)[1][2] (Kumar, et al., 2017), Google Actions (gActions)[3] powered by DialogFlow[4] for LU, Cortana Skills[5], Facebook Messenger Platform[6], and others (López, Quesada, & Guerrero, 2017). Each of these frameworks comes with developer-friendly web interface which allows developers to define their skills[7] and conversation flow. Speech recognition, LU, built-in intents, and predefined ready-to-use general entity types such as city names and airports come at no cost and require no integration effort.

However, developers are still required to: 1) provide sample utterances either as raw sentences or as templates with slots representing entity placeholders, 2) provide a dictionary of entity values for each domain-specific entity type, and 3) customize responses and interaction flows. Obviously, obtaining such domain-specific knowledge about entities of interest, and defining all possible structures of user utterances remain two key challenges. Moreover, developers have no control over the LU engine or its outcome, making interpretability and error analysis cumbersome. It is also unclear how the performance of these LU engines, in terms of accuracy, will scale when the task involves a large volume of (tens of) thousands of domain-specific entities, or how generalizable these engines are when user utterances involve new entities or new utterance structures.

---

[1] https://developer.amazon.com/alexa-skills-kit
[2] https://amzn.to/2qDjNcJ
[3] https://developers.google.com/actions/
[4] https://dialogflow.com/
[5] https://developer.microsoft.com/en-us/cortana
[6] https://developers.facebook.com/docs/messenger-platform/
[7] The task to be accomplished

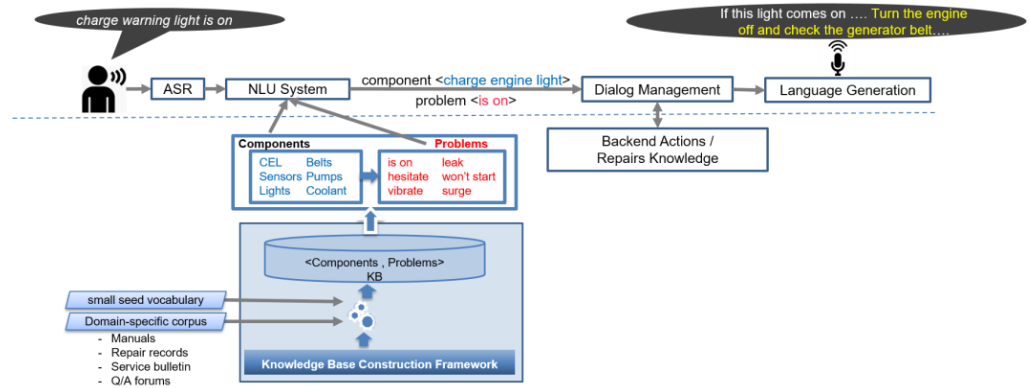

**Figure 1 High level architecture of our chatbot for responding to vehicle-related complaints**

In this paper, we describe the challenges and lessons learned from deploying a virtual assistant for suggesting repairs of equipment-related complaints. We demonstrate on two popular frameworks, namely ASK and gActions. Here, we focus on understanding and responding to vehicle-related problems, as an example equipment, which could be initiated by a driver or a technician. Along the paper, we try to answer three questions: 1) how can we facilitate the acquisition of domain-specific knowledge about entities related to equipment problems?; 2) how much knowledge off-the-shelf frameworks can digest effectively; 3) how accurately these frameworks built-in LU engines can identify entities in user utterances?.

Due to the scalability and accuracy limitations we experienced with ASK and gActions, we describe an alternative scalable pipeline for: 1) extracting the knowledge about equipment components and their associated problems entities, and 2) learning to identify such entities in user utterances. We show through evaluation on real dataset that the proposed framework understanding accuracy scales better with large volume of domain-specific entities being up to 30% more accurate.

## 2  Background and Related Work

Figure 1 shows the main components of our chatbot. In a nutshell, user utterance is firstly transcribed into text using the Automatic Speech Recognition (ASR) module. Then, the LU module identifies the entities (component, problem) in the input. Afterwards, the parsed input is passed to the dialog manager which keeps track of the conversation state and decides the next response (e.g., recommended repair) which is finally uttered back to the user using the language generation module.

As we mentioned earlier, we focus here on the cognitive-intensive task of creating the Knowledge Base (KB) of target entities on which the LU engine will be trained.

KB construction from text aims at converting the unstructured noisy textual data into a structured task-specific actionable knowledge that captures entities (elements of interest (EOI)), their attributes, and their relationships (Pujara & Singh, 2018). KBs are key components for many AI and knowledge-driven tasks such as question answering (Hao, et al., 2017), decision support systems (Dikaleh, Pape, Mistry, Felix, & Sheikh, 2018), recommender systems (Zhang, Yuan, Lian, Xie, & Ma, 2016), and others. KB construction has been an attractive research topic for decades resulting in many general KBs such as DBPedia (Auer, et al., 2007), Freebase (Bollacker, Evans, Paritosh, Sturge, & Taylor, 2008), Google Knowledge Vault (Dong, et al., 2014), ConceptNet (Speer & Havasi, 2013), NELL (Carlson, et al., 2010), YAGO (Hoffart, Suchanek, Berberich, & Weikum, 2013), and domain-specific KBs such as Amazon Product Graph, Microsoft Academic Graph (Sinha, et al., 2015).

The first step toward building such KBs is to extract information about target entities, attributes, and relationships between them. Several information extraction frameworks have been proposed in literature including OpenIE (Banko, Cafarella, Soderland, Broadhead, & Etzioni, 2007), DeepDive (Niu, Zhang, Ré, & Shavlik, 2012), Fonduer (Wu, et al., 2018), Microsoft QnA Maker (Shaikh, 2019), and others. Most of current information extraction systems utilize Natural Language Processing (NLP) techniques such as Part of Speech Tags (POS), shallow parsing, and dependency parse trees to extract linguistic features for recognizing entities.

Despite the extensive focus in the academic and industrial labs on constructing general purpose KBs, identifying component names and their associated problems in text has been lightly studied in literature.

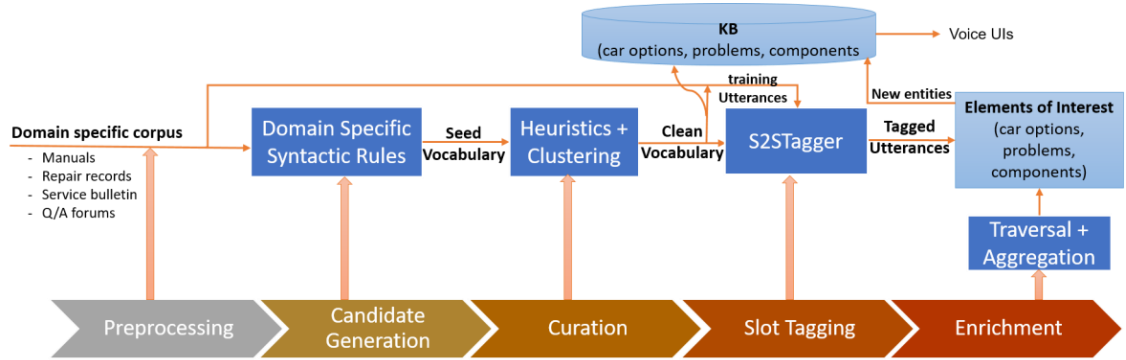

**Figure 2** The Knowledge base construction framework is a pipeline of five main stages

**Table 1** Sample vehicle complaint utterances (problems in red, and components in blue)

| Complaint Utterance |
| --- |
| low oil pressure |
| fuel filter is dirty |
| leak at oil pan |
| coolant reservoir cracked |
| pan leaking water |
| coolant tank is leaking |

The closest work to ours is (Niraula, Whyatt, & Kao, 2018) who proposed an approach to identify component names in service and maintenance logs using a combination of linguistic analysis and machine learning. The authors start with seed head nouns representing high level part names (e.g., valve, switch). Then extract all n-grams ending with these head nouns. Afterwards, the extracted n-grams are purified using heuristics. Finally, the purified part names are used to create an annotated training data for training a Conditional Random Fields (CRF) model (Lafferty, McCallum, & Pereira, 2001) to extract part names in raw sentences.

Similarly, (Chandramouli, et al., 2013) introduced a simple approach using n-gram extraction from service logs. Given a seed of part types, the authors extract all n-grams, with maximum of three tokens, which end with these part types. Then candidate n-grams are scored using a mutual information metric, and then purified using POS tagging.

Our framework automatically construct a KB of equipment components and their problems entities with "*component <has-a> problem*" relationships. Unlike previous work, we go one step further by extracting not only components and part names, but also their associated problems. Unlike (Niraula, Whyatt, & Kao, 2018), we start with syntactic rules rather than seed head nouns. The rules require less domain knowledge and should yield higher coverage. We then expand the constructed KB through two steps: 1) reorganizing the extracted vocabulary of components into a hierarchy using a simple traversal mechanism introducing *<is-a>* relationships (e.g., *stop light <is-a> light*), and 2) aggregating all the problems associated with subtype components in the hierarchy and associating them with supertype components introducing more *<has-a>* relationships (e.g., *coolant gauge <has-a> not reading → gauge <has-a> not reading*). Unsupervised curation and purification of extracted entities is another key differentiator of our framework compared to prior work. The proposed framework utilizes a state-of-the-art deep learning for sequence tagging to annotate raw sentences with component(s) and problem(s).

## 3  A Pipeline for KB Extraction

Existing chatbot development frameworks require knowledge about target entities[8] which would appear in users utterances. For each entity type (e.g., component, problem, etc.), an extensive vocabulary of possible values of such entities should be provided by the virtual assistant developer. These vocabularies are then used to train the underlying LU engine to identify entities in user utterance.

We propose a pipeline for creating a KB of entities related to vehicle complaint understanding from short texts, specifically posts in public Questions and Answers (QA) forums. Nevertheless, the design of the proposed framework is flexible and generic enough to be applied to several other maintenance scenarios of different equipment given a corpus with mentions of the same target entities. Table 1 shows sample complaint utterances from QA posts. As we can notice, most of these utterances are short sentences composed of a component along with an ongoing problem.

As shown in Figure 2, the proposed KB construction system is organized as a pipeline. We start with a domain-specific corpus that contains our target entities. We then process the corpus through five main stages including preprocessing, candidate generation using POS-based syntactic rules, embedding-based filtration and curation, and enrichment through training a sequence-to-sequence (seq2seq) slot tagging model. Our pipeline produces two outputs:

- A KB of three types of entities including car options (car, truck, vehicle, etc.), components, and their associated

---
[8] Slots in ASK terminology

Table 2 POS-based syntactic rules for candidate entity generation (problems in red, and components in blue)

| Utterance | POS | Rule |
|---|---|---|
| replace water pump | VB (NN\S*\s?)+ | (NN\S*\s?)+ → component |
| low oil pressure | JJ (NN\S*\s?)+ | JJ → problem, (NN\S*\s?)+ → component |
| fuel filter is dirty | (NN\S*\s?)+ VBZ JJ | (NN\S*\s?)+ → component, JJ → problem |
| coolant reservoir cracked | (NN\S*\s?)+ VBD | (NN\S*\s?)+ → component, VBD → problem |
| pan leaking water | (NN\S*\s?)+ VBG (NN\S*\s?)+ | (NN\S*\s?)+ → component, VBG (NN\S*\s?)+ → problem |
| coolant tank is leaking | (NN\S*\s?)+ VBZ VBG | (NN\S*\s?)+ → component, VBG → problem |

problems. These entities can be used to populate the vocabulary needed to build the voice-based agent in both ASK and DialogFlow.

- A tagging model which we call Sequence-to-Sequence Tagger (S2STagger). Besides its value in enriching the KB with new entities, S2Stagger can also be used as a standalone LU system that's able to extract target entities from raw user utterances.

In the following sub-sections, we will describe in more details each of the stages presented in Figure 2.

### 3.1 Preprocessing

Dealing with noisy text is challenging. In the case of equipment troubleshooting, service and repair records and QA posts include complaint, diagnosis, and correction text which represent highly rich resources of components and problems that might arise with each of them. Nevertheless, these records are typically written by technicians and operators who have time constraints and may lack language proficiency. Consequently, the text will be full of typos, spelling mistakes, inconsistent use of vocabulary, and domain-specific jargon and abbreviations. For these reasons, cautious use of preprocessing is required to reduce such inconsistencies and avoid inappropriate corrections. We perform the following preprocessing steps:

- Lowercase.
- Soft normalization: By removing punctuation characters separating single characters (e.g., a/c, a.c, a.c. → ac).
- Hard normalization: By collecting all frequent tokens that are prefixes of a larger token and manually replace them with their normalized version (e.g., temp → temperature, eng → engine, diag → diagnose…etc).
- Dictionary-based normalization: We create a dictionary of frequent abbreviations and use it to normalize tokens in the original text (e.g., chk, ch, ck → check)
- Manual tagging: We manually tag terms as vehicle, car, truck, etc. as a car-option entity.

### 3.2 Candidate Generation

To extract candidate entities, we define a set of syntactic rules based on POS tags of complaint utterances.

First, all sentences are extracted and parsed using the Stanford CoreNLP library (Manning, et al., 2014). Second, we employ linguistic heuristics to define chunks of tokens corresponding to component and problem entities based on their POS tags. Specifically, we define the rules considering only the most frequent POS patterns in our dataset.

Table 2 shows the rules defined for the most frequent six POS patterns. For example, whenever a sentence POS pattern matches an adjective followed by sequence of nouns of arbitrary length *(JJ (NN\S*\s?)+$)* (e.g. "*low air pressure*"), the adjective chunk is considered a candidate problem entity ("*low*") and the noun sequence chunk is considered a candidate component entity ("*air pressure*"). It is worth mentioning that, the defined heuristics are designed to capture components with long multi-term names which are common in our corpus (e.g., "*intake manifold air pressure sensor*"). We also discard irrelevant tokens in the extracted chunk such as determiners (a, an, the) preceding noun sequences and others.

### 3.3 Curation

In this stage, we prune incorrect and noisy candidate entities using weak supervision. We found that most of these wrong extractions were due to wrong annotations from the POS tagger due to the noisy nature of the text. For example, "*clean*" in "*clean tank*" was incorrectly tagged as adjective rather than verb causing "*clean*" to be added to the candidate problems pool. Another example "*squeals*" in "*belt squeals*" was tagged as plural noun rather than verb causing "*belt squeals*" to be added to the candidate components pool. To alleviate these issues, we employ different weak supervision methods to prune incorrectly extracted entities as follows:

- Statistical-based pruning: A simple pruning rule is to eliminate candidates that rarely appear in our corpus with frequency less than *F*.
- Linguistic-based pruning: These rules focus on the number and structure of tokens in the candidate entity. For example, a candidate entity cannot exceed *T* terms, must have terms with a minimum of *L* letters each, and cannot have alphanumeric tokens.

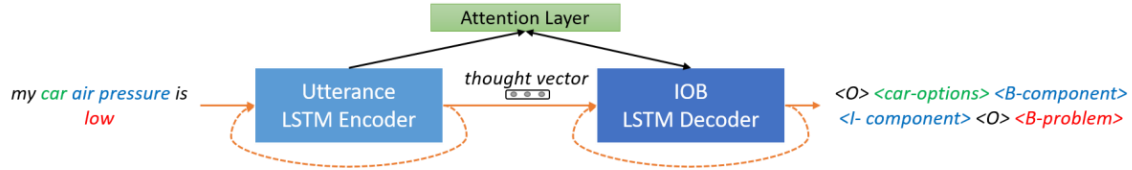

**Figure 3** S2STagger utilizes LSTM encoder-decoder to generate the IOB tags of input utterance. Attention layer is used to learn to softly align the input/output sequences

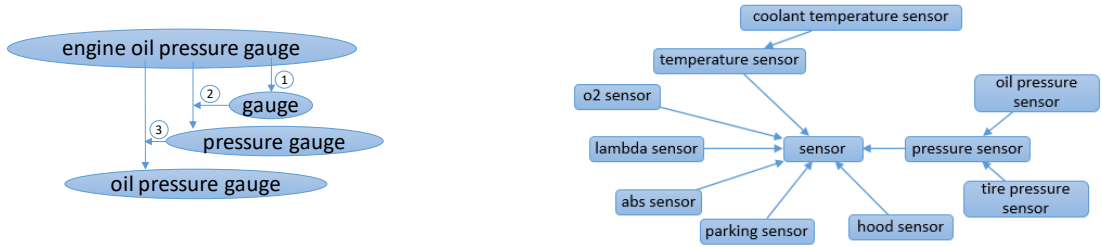

**Figure 4** Component hierarchy construction through backward traversal. Left – traversal through "*engine oil pressure gauge*" resulting in three higher level components. Right – example hierarchy with "*sensor*" as the root supertype component

- Embedding-based pruning: Fixed-length distributed representation models (aka embeddings) have proven effective for representing words and entities in many NLP tasks (Mikolov, Sutskever, Chen, Corrado, & Dean, 2013) (Shalaby, Zadrozny, & Jin, 2019). We exploit the fact that these models can effectively capture similarity and relatedness relationships between pairs of words and entities using their embeddings. To this end, we employ the model proposed by (Shalaby, Zadrozny, & Jin, 2019) to obtain the vector representations of all candidates. Then, we normalize all vectors and compute the similarity score between pairs of candidates using the dot product between their corresponding vectors. Afterwards, we prune all candidate problems that do not have at least *P* other problem entities with a minimum of $S_p$ similarity score. And prune all components that do not have at least *C* other component entities with a minimum of $S_c$ similarity score.
- Sentiment-based pruning: Utterances that express problems and issues usually have negative sentiment. With this assumption, we prune all candidate problem entities that are not semantically similar to at least one word from the list of negative sentiment words created by (Hu & Liu, 2004). Here, we measure the similarity score using the embeddings of candidate problem entities and the sentiment words as in the embedding-based pruning. Sentiment-based pruning helps discarding wrong extractions such as "*clean*" in "*clean tank*" where "clean" is tagged incorrectly as an adjective.

### 3.4 Slot tagging (S2STagger)

A desideratum of any information extraction system is to be lexical-agnostic; i.e., to be able to generalize well and identify unknown entities that have no mentions in the original dataset. Another desideratum is to be structure-agnostic; i.e., to be able to generalize well and identify seen or new entities in utterances with different structures from those in the original dataset. Rule-based candidate extraction typically yields highly precise extractions.

However, depending solely on rules limits the system recognition capacity to mentions in structures that match these predefined rules. Moreover, it is infeasible to handcraft rules that cover all possible complaint structures limiting the system recall. It is also expected that new components and problems will emerge, especially in highly dynamic domains, and running the rules on an updated snapshot of the corpus would be an expensive solution.

A more practical and efficient solution is to build a machine learning model to tag raw sentences and identify chunks of tokens that correspond to our target entities. To this end, we adopt a neural attention-based seq2seq model called S2STagger to tag raw sentences and extract target entities from them. To train S2STagger, we create a dataset from utterances that match our syntactic rules and label terms in these utterances using the inside-outside-beginning (IOB) notation (Ramshaw & Marcus, 1999). For example, "*the car air pressure is low*" would be tagged as "*<O> <car-options> <B-component> <I- component> <O> <B-problem>*". As the extractions from the syntactic rules followed by curation are highly accurate, we expect to have highly accurate training data for our tagging model. It is worth mentioning that we only use utterances with mentions of entities not pruned during the curation phase.

As shown in Figure 3, S2STagger utilizes an encoder-decoder Recurrent Neural Network architecture (RNN) with Long-Short Term Memory (LSTM) cells (Gers, Schmidhuber, & Cummins, 1999). During encoding, raw terms in each sentence are processed sequentially through an RNN and encoded as a fixed-length vector that captures all the semantic and syntactic structures in the sentence. Then, a decoder RNN takes this vector and produces a sequence of IOB tags, one for each term in the input sentence. Because each tag might depend on one or more terms in the input but not the others, we utilize an attention mechanism so that the network learns what terms in the input are more relevant for each tag in the output (Bahdanau, Cho, & Bengio, 2014) (Luong, Pham, & Manning, 2015).

**Table 3** Dataset Statistics

|  | **Mechanics StackExchange + Yahoo QA** |
|---|---|
| Sample utterances | • rough start<br>• battery drains when AC is used |
| # of utterances | 574,432 |
| # of utterances matching syntactic-rule | 11,619 (~2%) |
| # of extracted components | 5,972 |
| # of extracted problems | 2,455 |

### 3.5 KB Consolidation and Enrichment

At this stage, we enrich the KB with new entities not explicitly mentioned in the training utterances. These new entities are obtained from three different sources:

- S2STagger: After training S2STagger, we use it to tag the remaining utterances in our dataset which do not match our syntactic rules, resulting in a new set of entities. Importantly, the trained S2Stagger model can be used to tag newly unseen utterances allowing the proposed KB framework to scale efficiently whenever new utterances are collected.
- Component backward traversal: We propose using a simple traversal method to create a hierarchy of supertype components from the extracted components vocabulary after curation and tagging using S2STagger. As shown in Figure 4, we consider each extracted component (subtype) and backward traverse through its tokens one token at a time. At each step, we append the new token to the component identified in the previous traversal step (supertype). For example, traversing "*engine oil pressure gauge*" will result in "*gauge*", "*pressure gauge*", and "*oil pressure gauge*" in order. As we can notice, components at the top of the hierarchy represent high level and generic ones (supertypes) which can be common across domains (e.g., *sensor, switch, pump, etc.*). The hierarchy allows introducing "*subtype <is-a> supertype*" relationship between components enriching the KB with more supertype components.
- Problem aggregation: The new components identified through backward traversal will initially have no problems associated with them. We propose using a simple aggregation method to automatically associate between supertype components and problems of their subtypes. First, we start with the leaf subtype components in the hierarchy. Second, we navigate through the hierarchy upward one level at a time. At each step, we combine all the problems from the previous level and associate them to the supertype component at the current level. For example, all problems associated with "*oil pressure sensor*", "*tire pressure sensor*", etc. will be aggregated and associated with "pressure sensor". Then problems of "*pressure sensor*", "*o2 sensor*", "*abs sensor*", etc. will be aggregated and associated with "sensor". This simple aggregation method allows introducing new "*supertype <has-a> problem*" relationships in the constructed KB.

We consolidate the entities from the curation stage along with the new entities discovered at each of the three steps to create our KB of components and problems entities as shown in Figure 2.

## 4 Data and Model Evaluation

### 4.1 Dataset

We experiment our framework with two datasets in the automotive sector. First, a dataset of Questions and Answers (QA) from the public Mechanics Stack Exchange[9] QA forum. Second, another subset of questions related to cars maintenance from the Yahoo QA dataset. Table 3 shows some example utterances and statistics from the datasets. As we can see, the coverage of syntactic rules is noticeably low. This demonstrates the need for a learning-based entity extractor, such as our proposed S2STagger model to harness the knowledge from utterances not matching the predefined rules.

### 4.2 Augmentation and Model Training

We create a labeled dataset for S2STagger using the question utterances. After candidate extraction, we curate the extracted entities using F=1, T=6 and 5 for components and problems respectively, L=2 and 2 for components and problems recursively, P=1, $S_p$=0.5, C=5, $S_c$=0.5. Then, we only tag utterances with entity mentions found in the curated entities pool.

### 4.3 Quantitative Evaluation

One of the main pain points in creating voice-enabled agents is the definition of the vocabulary users will use to communicate with the system. This vocabulary is typically composed of large number of entities which are very costly and time consuming to define manually. Our framework greatly facilitates this step through the construction of KBs with target entities which could be readily available to build voice-enabled agents. To assess the effectiveness of the extracted knowledge, we create three tagging models. Two models using off-the-shelf NLU technologies including: 1) a skill (AlexaSkill) using Amazon Alexa Skills kit[10], and 2) an agent (DiagFlow) using Google Dialog Flow[11]. With both models, we define utterance structures equivalent to the syntactic rules structures. We also feed the same curated entities in our KB to both models as slot values and entities for AlexaSkill and DiagFlow respectively. The third model is S2STagger trained on all the tagged utterances in the QA dataset. It's important to emphasize that, the training utterances for S2STagger are the same from which KB entities and utterance structures are extracted and fed to both AlexaSkill and DiagFlow. Due to model size limitations imposed by these frameworks, we couldn't feed the raw utterances to both agents as we did with S2STagger.

---

[9] https://mechanics.stackexchange.com/
[10] https://developer.amazon.com/alexa
[11] https://dialogflow.com/

Table 4 Evaluation dataset including vehicle-related complaints

|  | Utterances |
|---|---|
| Same structure/Same entities | 119 |
| Same structure/Different entities | 75 |
| Different structure/Same entities | 20 |
| **All** | **214** |

Table 5 Accuracy on vehicle-related complaints dataset

|  | Accuracy (Exact IOB Match) | | |
|---|---|---|---|
|  | AlexaSkill | DiagFlow | S2STagger |
| Same structure/Same entities (119) | (83) 70% | (47) 39% | **(111) 93%** |
| Same structure/Different entities (75) | (40) 53% | (7) 9% | **(67) 89%** |
| All (194) | (123) 63% | (54) 28% | **(178) 92%** |

Table 6 Accuracy on utterances with different structure / same entities

|  | Accuracy | | |
|---|---|---|---|
|  | (Exact IOB Match) | component entities (20) | problem entities (19) |
| AlexaSkill | 0% | 0% | 0% |
| DiagFlow | 0% | 4 (45%) | 0% |
| S2STagger | (2) 1% | **9 (45%)** | **3 (16%)** |

We create an evaluation dataset of utterances that were manually tagged. The dataset describes vehicle-related complaints and shown in Table 4. The utterances are chosen such that three aspects of the model are assessed. Specifically, we would like to quantitatively measure the model accuracy on utterances with: 1) same syntactic structures and same entities as in the training utterances (119 in total), 2) same syntactic structures but different entities from the training utterances (75 in total), and 3) different syntactic structures but same entities as in the training utterances (20 in total). It is worth mentioning that, to alleviate the out-of-vocabulary (OOV) problem, different entities are created from terms in the model vocabulary. This way, incorrect tagging can only be attributed to the model inability to generalize to entities tagged differently from the training ones.

Table 5 shows that accuracy of S2STagger compared to the other models on the car complaints evaluation dataset. We report the exact match accuracy in which each and every term in the utterance must be tagged same as in ground truth for the utterance to be considered correctly tagged. As we can notice, S2STagger gives the best exact accuracy outperforming the other models significantly. Interestingly, with the utterances that have same structure but different entities, S2STagger tagging accuracy is close to same structure / same entities utterances. This indicates that our model is more lexical-agnostic and can generalize better than the other two models. AlexaSkill model comes second, while DiagFlow model can only tag few utterances correct, indicating its heavy dependence on exact entity matching and limited generalization ability.

On the other hand, the three models seem more sensitive to variation in utterance syntactic structure than its lexical variation. As we can notice in Table 6, the three models fail to correctly tag almost all the utterances with different structure (S2STagger tags 2 of 20 correct). Even when we measure the accuracy on the entity extraction level not the whole utterance, both AlexaSkill and DiagFlow models still struggle with understanding the different structure utterances. S2Stagger, on the other hand, can tag 9 component entities and 4 problem entities correctly, which is still lower than its accuracy on the utterances with same structure.

**Table 7** Success and failure tagging of the three models. Bold indicates incorrect tagging

| Utterance | | |
|---|---|---|
| **Same structure/Same entities** | | |
| my car steering wheel wobbles | Ground Truth | <O> <car-option> <B-component> <I-component> <B-problem> |
| | AlexaSkill | <O> <car-option> <B-component> <I-component> <B-problem> |
| | DiagFlow | <O> <car-option> <B-component> <I-component> <B-problem> |
| | S2STagger | <O> <car-option> <B-component> <I-component> <B-problem> |
| car has low coolant | Ground Truth | <car-option> <O> <B-problem> <B-component> |
| | AlexaSkill | <car-option> <O> <B-problem> <B-component> |
| | DiagFlow | **fail** |
| | S2STagger | <car-option> <O> <B-problem> <B-component> |
| **Same structure/different entities** | | |
| clutch pedal is hard to push | Ground Truth | <B-component> <I-component> <O> <B-problem> <I-problem> <I-problem> |
| | AlexaSkill | <B-component> <I-component> <O> <B-problem> <I-problem> <I-problem> |
| | DiagFlow | **fail** |
| | S2STagger | <B-component> <I-component> <O> <B-problem> <I-problem> <I-problem> |
| wrapped brake rotor | Ground Truth | <B-problem> <B-component> <I-component> |
| | AlexaSkill | **fail** |
| | DiagFlow | **fail** |
| | S2STagger | <B-problem> <B-component> <I-component> |
| **Different structure/same entities** | | |
| the steering wheel in my car wobbles | Ground Truth | <O> <B-component> <I-component> <O> <O> <car-option> <B-problem> |
| | AlexaSkill | **fail** |
| | DiagFlow | **fail** |
| | S2STagger | **<O> <B-problem> <B-component> <I-component> <O> <car-option> <B-problem>** |
| low car coolant | Ground Truth | <B-problem> <car-option> <B-component> |
| | AlexaSkill | **<B-problem> <O> <B-component>** |
| | DiagFlow | **fail** |
| | S2STagger | <B-problem> <car-option> <B-component> |

## 4.4 Qualitative Evaluation

To better understand these results, we show some examples of success and failure cases for the three models in Table 7. As we can notice, the models work well on what it has seen before (same structure and same entities). On the other hand, when the same entities appear in paraphrased version of the training utterance (e.g., *"car has low coolant"* vs. *"low car coolant"*), the models generally fail to recognize them. When it comes to generalizing to different from training entity mentions in utterances with same structures such as *"hard to push"* and *"brake rotor"*, S2STagger generalizes better than the two other models though *"hard"* and *"brake"* were already labeled as entities in the training data.

More importantly, even though S2STagger can successfully tag new entities (*"hard to push"* and *"brake rotor"*) when they appear in similar to training structures; it fails to recognize same entities when they appear in slightly different structure (e.g., *"steering wheel in my car wobbles"* vs. *"car steering wheel wobble"*). These examples demonstrate the ability of S2STagger to generalize to unseen entities better than unseen structures. It also demonstrates how the other two models seem to depend heavily to lexical matching causing them to fail to recognize mentions of new entities.

## 5 Conclusion

The proposed pipeline serves our goal toward automatically constructing the knowledge required to understand equipment-related complaints in arbitrary target domain (e.g., vehicles, appliances, etc.). By creating such knowledge about components and problems associated with them, it is possible to identify what the user is complaining about.

One of the benefits of the proposed KB construction framework is facilitating the development and deployment of intelligent conversational assistants for various industrial AI scenarios (e.g., maintenance & repair, operations, etc.) through better understanding of user utterances. As we demonstrated in section 4, the constructed KB from QA forums facilitated developing two voice-enabled assistants using ASK and DialogFlow without any finetuning or adaption to either vendors. Thus, the proposed pipeline is an important tool to potentially automate the deployment process of voice-enabled AI solutions making it easy to use NLU systems of any vendor. In addition, S2STagger provides scalable and efficient mechanism to extend the constructed KB beyond the handcrafted candidate extraction rules.

Another benefit of this research is to improve existing maintenance & repair solutions through better processing of user complaint text. In other words, identifying components(s) and problem(s) in the noisy user complaints text and focusing on these entities only while predicting the repair.

The results demonstrate superior performance of the proposed knowledge construction pipeline including S2STagger, the slot tagging model, over popular systems such as ASK and DialogFlow in understanding vehicle-related complaints. One important and must have feature is to increase the effectiveness of S2STagger and off-the-shelf NLU systems to handle utterances with different from training structures. We think augmenting the training data with carrier phrases is one approach. Additionally, training the model to paraphrase and tag jointly could be a more genuine approach as it does not require to manually define the paraphrasing or carrier phrases patterns.

There were also some of the issues that impacted the development of this research. For example, the limited scalability of off-the-shelf NLU systems: ASK model size cannot exceed 1.5MB, while DialogFlow Agents cannot contain more than 10K different entities. Deployment of the constructed KB on any of these platforms would be limited to a subset of the extracted knowledge. Therefore, it seems mandatory for businesses and R&D labs to develop in-house NLU technologies to bypass such limitations.